\documentclass[letterpaper, 10 pt, conference]{ieeeconf}
\usepackage{amsfonts}
\usepackage{eurosym}
\usepackage{graphics}
\usepackage{epsfig}
\usepackage{amsmath}
\usepackage{amssymb}
\usepackage{mathrsfs}
\usepackage[ruled,vlined]{algorithm2e}
\usepackage{xcolor}
\usepackage{subcaption}
\DeclareMathOperator*{\argmin}{arg\,min}

\IEEEoverridecommandlockouts
\newtheorem{theorem}{Theorem}

\newtheorem{definition}{Definition}

\newtheorem{lemma}{Lemma}

\newtheorem{assumption}{Assumption}
\input{tcilatex}

\begin{document}

\title{{\LARGE \textbf{Active Information Acquisition under Arbitrary Unknown Disturbances}}}
\author{Jennifer Wakulicz, He Kong\thanks{%
The authors are with the Australian Centre for Field Robotics, The
University of Sydney, NSW, 2006, Australia. Emails:
jwak2935@uni.sydney.edu.au, he.kong@sydney.edu.au,
salah.sukkarieh@sydney.edu.au. Corresponding author: He Kong.}, and Salah
Sukkarieh}
\maketitle

\begin{abstract}Trajectory optimization of sensing robots to actively gather information of targets has received much attention in the past. It is well-known that under the assumption of linear Gaussian target dynamics and sensor models the stochastic Active Information Acquisition problem is equivalent to a deterministic optimal control problem. However, the above-mentioned assumptions regarding the target dynamic model are limiting. In real-world scenarios, the target may be subject to disturbances whose models or statistical properties are hard or impossible to obtain. Typical scenarios include abrupt maneuvers, jumping disturbances due to interactions with the environment, anomalous misbehaviors due to system faults/attacks, etc. Motivated by the above considerations, in this paper we consider targets whose dynamic models are subject to arbitrary unknown inputs whose models or statistical properties are not assumed to be available. In particular, with the aid of an unknown input decoupled filter, we formulate the sensor trajectory planning problem to track evolution of the target state and analyse the resulting performance for both the state and unknown input evolution tracking. Inspired by concepts of Reduced Value Iteration, a suboptimal solution that expands a search tree via Forward Value Iteration with informativeness-based pruning is proposed. Concrete suboptimality performance guarantees for tracking both the state and the unknown input are established. Numerical simulations of a target tracking example are presented to compare the proposed solution with a greedy policy.
\end{abstract}

\thispagestyle{empty} \pagestyle{empty}



\section{Introduction}

Due to its vast applications such as environmental monitoring \cite{Singh2009}, target/source motion tracking/localization \cite{EiffertICRA}-\cite{KongTRO}, agriculture \cite{EiffertCASE}-\cite{ISER2021}, sensor management has been studied extensively in robotics and automation literature, in terms of communication management \cite{Sukkarieh2014}-\cite{Sukkarieh2015} and sensor trajectory planning \cite{Pappas2009}-\cite{RSS2019}, etc. Closely related problems of sensor scheduling and sensor placement have also received much attention in the control community \cite{Tomlin2012}-\cite{Mo2021}.

The problem of managing one or more sensor-equipped mobile robots' trajectories to maximize the information gathered regarding a target system/process is known as Active Information Acquisition (AIA). The AIA problem is commonly formulated as a stochastic control problem where the mutual information between sensor measurements and the target state is optimized \cite{Atanasov2014}. When the target's motion dynamics is linear and driven only by Gaussian noise, and the sensor's observation model is linear in the target state, it is well-known that the stochastic AIA problem reduces to a deterministic optimal control problem for which open-loop solutions are optimal (see \cite{Pappas2009} and the references therein). Recent works have also established tree-search based methods and algorithms to efficiently approximate the optimal policy while maintaining suboptimality guarantees \cite{Atanasov2014}, \cite{Tomlin2012}.

However, the assumptions made regarding the target motion model in the aforementioned works are often limiting. For example, the target may be subject to arbitrary unknown disturbances which are difficult, if possible, to statistically interpret or model. Typical examples in applications include systems under fault/attacks  \cite{Frazzoli2016}-\cite{Tabuada}, tracking/localisation of targets subject to abrupt maneuvers \cite{Dixon2011}-\cite{Bernstein2019}, advanced vehicle applications under complex tire-ground interactions \cite{Khajepour2017}-\cite{Cao2019}, estimation of unmeasured forces in grasping/manipulation \cite{Kang2012}-\cite{Ritter2020}, etc. In fact, filtering and estimation under arbitrary unknown inputs have received much attention in the control literature \cite{Gillijns2007A}-\cite{Kong2020Auto}, and have found numerous applications, including robustness/security analysis and synthesis of resilient autonomous robots and connected vehicle systems \cite{Zhu2018}-\cite{Trans2019}.

Motivated by the above considerations, here we consider targets whose dynamics are subject to arbitrary unknown disturbances. In this case, it is of interest to track the evolution of both the target and the unknown disturbances. To circumvent complexities of such an approach, we formulate and solve the AIA problem for tracking the target state and analyse the resulting performance of both target state and unknown disturbance tracking. Firstly, we show that both the state and input error covariance update maps given in existing unknown input filtering works \cite{Gillijns2007A}-\cite{Kong2020Auto} are concave and monotone. To the best of our knowledge, these properties have not previously been explored. Secondly, inspired by the concepts of Reduced Value Iteration (RVI) \cite{Atanasov2014},\cite{Tomlin2012}, we propose a suboptimal solution to the AIA problem using Forward Value Iteration (FVI) with pruning according to an information dominance metric. Concrete suboptimality performance guarantees for tracking both the target state and the unknown disturbance are established. Finally, we use a target tracking example to show the merits of the proposed solution in comparison to a greedy policy.


\section{\label{prelim}Preliminaries and Problem Formulation}

\subsection{\label{UIF}Preliminaries of filtering under unknown inputs}

Consider a mobile sensor with discrete dynamics model:%
\begin{equation}
x_{k+1}=f(x_{k},u_{k}),  \label{sensor_motion}
\end{equation}%
where $x_{k}\in \mathcal{X}\cong \mathbf{R}^{n_{x}}$ is the sensor state
with $\mathcal{X}$ being an $n_{x}-$dimensional state space with metric $d_{%
\mathcal{X}}$, and $u_{k}\in \mathcal{U}$ is the control input with $\mathcal{U}$
as a finite space of admissible controls. Suppose there exists a target with linear time-varying motion model:%
\begin{equation}
y_{k+1}=A_{k}y_{k}+G_{k}d_{k}+w_{k},  \label{target_model}
\end{equation}%
where $y_{k}\in \mathbf{R}^{n_{y}}$ is the target state vector, the target
process noise $w_{k}\sim \mathcal{N}(0,Q_{k})$, i.e. $w_{k}$ is normally
distributed with zero-mean and covariance $Q_{k}\in \mathbf{R}^{n_{y}\times
n_{y}}\succ0$, $d_{k}\in \mathbf{R}^{n_d}$ represents arbitrary unknown inputs
whose models or statistical properties are not assumed to be known and $%
A_{k},G_{k}$ are known matrices of compatible dimensions. Without loss of
generality, we assume that $rank(G_{k})=n_d.$ While in operation, the sensor
has observation model:%
\begin{equation}
z_{k}=C_{k}(x_k)y_{k}+v_{k}(x_k),  \label{sensor_mea}
\end{equation}%
where, $z_{k}\in \mathbf{R}^{n_z}$ is the measurement, $C_{k}(x_k)\in \mathbf{R}^{n_z\times n_{y}}$ is a known measurement matrix, and the
measurement noise $v_{k}(x_k)\sim \mathcal{N}(0,R_{k}(x_k))$ with $R_{k}(x_k)\succ0$. For brevity we drop dependence of $C_k$, $R_k$, $v_k$ on the sensor state $x_k$ in the remainder of the paper.

For filtering purposes, we adopt the
framework of \cite{Gillijns2007A} (other methods in \cite{Frazzoli2016}, \cite{Kong2020Auto} can be similarly considered) and implement the following steps
recursively after initialization:
\begin{enumerate}
\item[\textbf{1.}] Time update: 
\begin{equation}
\hat{y}_{k|k-1}=A_{k}\hat{y}_{k-1|k-1},  \label{pre_without}
\end{equation}

\item[\textbf{2.}] Unknown input estimation:%
\begin{equation}
\hat{d}_{k-1}=M_{k}(z_{k}-C_{k}\hat{y}_{k|k-1}), \text{ }M_{k}\in \mathbf{%
R}^{n_d\times n_z},  \label{d_est}
\end{equation}

\item[\textbf{3.}] Measurement update:%
\begin{equation}\label{unbias_x_est}%
\begin{array}{ll}
\hspace{-3mm}\hat{y}_{k|k}^{\star} = \hat{y}_{k|k-1}+G_{k-1}\hat{d}_{k-1}\\
\hspace{-3mm}\hat{y}_{k|k} = \hat{y}_{k|k}^{\star }+K_{k}(z_{k}-C_{k}\hat{y}_{k|k}^{\star }),  K_{k}\in \mathbf{R}^{n_{y}\times n_z}.%
\end{array}%
\end{equation}%
\end{enumerate}%
Define%
\begin{equation}
\begin{array}{l}
\widetilde{d}_{k-1}=d_{k-1}-\widehat{d}_{k-1}, \hspace{5mm}\Sigma_{k-1}^{d}=\mathbb{E}[%
\widetilde{d}_{k-1}\widetilde{d}_{k-1}^{\mathrm{T}}], \\ 
\widetilde{y}_{k\mid k}=y_{k}-\hat{y}_{k|k},\hspace{5mm}\Sigma _{k}=\mathbb{E}[%
\widetilde{y}_{k\mid k}\widetilde{y}_{k\mid k}^{\mathrm{T}}],%
\end{array}
\label{definition_error}
\end{equation}%
as the unknown input estimation error, the filtered state error, and their respective covariances.

As shown in \cite{Gillijns2007A}, $\hat{d}_{k-1}$ and $\hat{y}_{k|k}$ in (\ref{d_est})-(\ref%
{unbias_x_est}) are unbiased estimates if and only if the initial state guess $\hat{y}_{0|0}$ is
unbiased and the unknown input filter gain $M_{k}$ satisfies%
\begin{equation}\label{gain_constraint}
M_{k}C_{k}G_{k-1}=I_{n_{d}}.  
\end{equation}%
The \textit{optimal} unknown input filter gain in the minimum variance
sense is given by%
\begin{equation}
M_{k}^*(\Sigma_{k-1})=(F_{k}^{\mathrm{T}}\widetilde{R}_{k}^{-1}(\Sigma_{k-1})F_{k})^{-1}F_{k}^{\mathrm{T}%
}\widetilde{R}_{k}^{-1}(\Sigma_{k-1}),  \label{M_k}
\end{equation}%
where $F_{k}=C_{k}G_{k-1}$,  $\widetilde{R}_{k}(\Sigma_{k-1})=C_{k}(A_{k-1}\Sigma
_{k-1}A_{k-1}^{\mathrm{T}}+Q_{k-1})C_{k}^{\mathrm{T}}+R_{k}\succ0$ and $\Sigma _{k-1}$ is the filtered state error covariance at time step $%
k-1$. Given $M_k^*$, one may transform the state estimation problem into a standard Kalman filtering problem and find a resulting optimal gain matrix $K_k^*$ \cite{Gillijns2007A}. The resulting optimal gain $K_k^*$ is in general non-unique \cite{Gillijns2007A}. For simplicity, in this paper we take the choice
\begin{equation}\label{state_gain}
    K_k^*(\Sigma_{k-1}) = (A_{k-1}\Sigma_{k-1}A_{k-1}^{\mathrm{T}}+Q_{k-1})C_k^{\mathrm{T}}\widetilde{R}_k^{-1}.
\end{equation}

The optimal filter gains in (\ref{M_k})-(\ref{state_gain}) give the state and unknown input error covariance update maps respectively:%
\begin{equation}\label{neat_update}
\begin{array}{l}
\Sigma_{k} =  \rho (\Sigma _{k-1},M^{*}_{k},K^{*}_{k}) \\ = \widetilde{A}_{k}\Sigma _{k-1}\widetilde{A}_{k}^{\mathrm{T}}+\widetilde{F}_{k}Q_{k-1}\widetilde{F}_{k}^{\mathrm{T}}+\widetilde{W}_{k}R_{k}\widetilde{W}_{k}^{\mathrm{T}},\\
\Sigma _{k-1}^{d}=\rho^{d}(\Sigma _{k-1}) = (F_{k}^{\mathrm{T}}\widetilde{R}_{k}^{-1}(\Sigma_{k-1})F_{k})^{-1},%
\end{array}
\end{equation}
where
\begin{equation*}
\begin{array}{l}
\widetilde{A}_{k}=(I-K^{*}_{k}C_{k})(I-G_{k-1}M^{*}_{k}C_{k})A_{k-1}, \\ 
\widetilde{F}_{k}=-(I-K^{*}_{k}C_{k})(I-G_{k-1}M^{*}_{k}C_{k}), \\ 
\widetilde{W}_{k}=G_{k-1}M^{*}_{k}-K_{k}C_{k}G_{k-1}M^{*}_{k}+K^{*}_{k}.
\end{array}
\end{equation*}
Note that $\rho$, $\rho^d$ are indeed functions of $x_k$ and thus of the control input $u_{k-1}$. We therefore denote $\rho_{u_k}(\Sigma_k)$, $\rho_{u_k}^d(\Sigma_k)$ to refer to the update maps applied under the control $u_k\in\mathcal{U}$.

\subsection{Problem Formulation\label{prob_form}}

Given an initial sensor state $x_{0}\in \mathcal{X}$ and a prior
distribution of the target state $y_{0},$ the problem of interest is to
optimize the trajectory of the sensor over a planning horizon of length $N$ to best track the evolution of the target dynamics and the unknown input. Expanding upon the problem formulation in \cite{Atanasov2014}, \cite{Tomlin2012}, we consider the following
optimal control problem
\begin{equation}\label{problem1}
\begin{array}{l}
\min\limits_{\sigma\in \mathcal{U}^{N}}\log \det (\Sigma_{N})+\log \det
(\Sigma _{N-1}^{d}) 
\end{array}
\end{equation}
\begin{equation*}
\begin{array}{lll}s.t.&x_{k+1}=f(x_{k},u_{k}), &k=0,\ldots ,N-1, \\ 
&\Sigma _{k+1}=\rho_{u_k}(\Sigma _{k}), &k=0,\ldots ,N-1, \\ 
&\Sigma _{k}^{d}=\rho_{u_k}^{d}(\Sigma _{k}),&k=1,\ldots ,N-1,
\end{array}
\end{equation*}
where $\sigma=\left\{ u_{0},\cdots ,u_{N-1}\right\} \in \mathcal{U}^{N}$ is a sequence of admissible controls, and $\rho_{u_k}
(\Sigma _{k})$, $\rho_{u_k}^{d}(\Sigma _{k})$ are the state and unknown input error covariance update maps defined in (\ref{neat_update}), with the first measurement taken at sampling instant $k=1$.




Although the unknown inputs are not assumed to follow any
specific probability distribution, one could give some statistical
interpretation of the optimization problem (\ref{problem1}) similar to the
existing works for the case without unknown inputs \cite{Pappas2009}, \cite{Atanasov2014}. This can be done by following the concepts in \cite%
{Bitmead2019} to firstly pose the unknown input as a Gaussian noise process
with variance $D$ and derive the statistical interpretation of problem (\ref{problem1}). Then, the lack of prior information regarding the unknown input can be expressed by taking $D$ to infinity. Due to limited space, we will not pursue this
point further.



Finding the optimal solution to problem (\ref{problem1}) amounts to exploring the large space of sensor states and error covariances allowed by $\mathcal{U}$ over a planning horizon $N$ and finding the optimal path via tree search.
To obtain a compromise between complexity and optimality of search tree construction, we adopt the concepts of the RVI algorithm proposed in \cite{Atanasov2014}, \cite{Tomlin2012}. Conceptually, if a set of nodes are sufficiently close in sensor configuration space (i.e. they $\delta$-cross) and one node's covariance is not as informative as nearby nodes' (i.e. is $\epsilon$-algebraically redundant), it is discarded from the tree. This method reduces computational complexity and gives suboptimality bounds for the resulting solution. $\delta$-crossing and $\epsilon$-algebraic redundancy are formalized below.
\begin{definition}\cite{Atanasov2014}\label{d_cross} Two sensor trajectories $\delta$-cross at time $k\in [1,N]$ if  $d_{\mathcal{X}}(x_k^1, x_k^2)\leq\delta,$ for $\delta\geq 0$.
\end{definition}
\begin{definition}\cite{Tomlin2012}\label{alg_redun} Let $\epsilon \geq 0$ and $\{\Sigma^i\}_{i=1}^{K}$ be a finite set with $\Sigma^i \succeq 0$ $\forall i$. Then a matrix $\Sigma \succeq 0$ is $\epsilon$-algebraically redundant with respect to $\{\Sigma^i\}_{i=1}^{K}$ if there exists a set of nonnegative constants $\{\alpha_i\}_{i=1}^{K}$ such that
\begin{equation*}
\begin{array}{ll}
     \sum_{i=1}^K \alpha_i = 1, & \Sigma + \epsilon I \succeq \sum_{i=1}^K\alpha_i\Sigma^i.
\end{array}
\end{equation*}
\end{definition}

To prune nodes $(x_{k}, \Sigma_{k}, \Sigma_{k-1}^d)$ according to their algebraic redundancy and approximately solve (\ref{problem1}) one must consider how informative $\delta$-crossing nodes are for state evolution and unknown input evolution tracking separately. That is, Definition \ref{alg_redun} must be checked for both $\Sigma_k$ and $\Sigma_{k-1}^d$. This may result in highly informative nodes for state evolution tracking being pruned due to mediocre contribution to unknown input tracking, making suboptimality bounds for the resulting solution to (\ref{problem1}) difficult to analyse. 

However, the close relationship between state and unknown input error covariance update maps seen in (\ref{neat_update}) allows one to prune according to state tracking performance only while still having concrete performance guarantees for unknown input tracking. The following sections of the paper therefore address the reduced problem
\begin{equation}\label{problem_reduce}
\begin{array}{l}
\min\limits_{\sigma\in \mathcal{U}^{N}}\log \det (\Sigma_{N})
\end{array}
\end{equation}
\begin{equation*}
\begin{array}{lll}s.t.&x_{k+1}=f(x_{k},u_{k}), &k=0,\ldots ,N-1, \\ 
&\Sigma _{k+1}=\rho_{u_k}(\Sigma _{k}), &k=0,\ldots ,N-1,
\end{array}
\end{equation*}
and derive suboptimality bounds for both state and unknown input tracking that result from this simplified approach.

The RVI algorithm for tracking targets with unknown dynamics is detailed in Algorithm \ref{RVI_pseudocode}. Note that in Algorithm \ref{RVI_pseudocode}, we use the reduced cost function in (\ref{problem_reduce}). The expansion of nodes in Line $5$ is performed using the filter introduced in (\ref{pre_without})-(\ref{neat_update}) to accommodate for any non-zero $d_k$.

\SetKwFunction{checkRedundancy}{checkRedundant}
\SetKwFunction{False}{False}
\SetKw{Or}{or}
\LinesNumbered
\begin{algorithm}[t]
\SetAlgoVlined
Initialise $S_k = \emptyset$ for $k\in[1,N]$, $S_0 = {(x_0,\Sigma_0)}$\;
\ForAll{$k \in [1,N]$}{
    \ForAll{$(x,\Sigma,\Sigma^d)\in S_{k-1}$}{
        \ForAll{$u \in \mathcal{U}$}{
        $S_k \leftarrow S_k \bigcup \{(f(x,u),\rho_u(\Sigma),\rho_u^d(\Sigma)\}$\;
        }
    }
    $S_{min} \leftarrow \{(x,\Sigma,\Sigma^d)\in S_k \mid \Sigma = \argmin(\log \det(\Sigma_k)) \}$\; 
    $S_k' \leftarrow S_{min}$\; 
    \ForAll{$(x,\Sigma,\Sigma^d)\in S_k\setminus S_{min}$}{
        $Q \leftarrow \{\Sigma \mid (x', \Sigma',\Sigma^{d'}) \in S_k', d_{\mathcal{X}}(x',x) \leq \delta\}$\;
        \If{$Q = \emptyset$ \Or{\checkRedundancy{$\Sigma,Q$} is \False}}{
            $S_k' \leftarrow S_k'\bigcup(x,\Sigma,\Sigma^d)$\;
        }
    }
}\KwRet{$\Sigma_{N}^{\epsilon,\delta} = \argmin \left(\log\det(\Sigma_N)\right)$\;}
\caption{RVI With Unknown Inputs \label{RVI_pseudocode}}
\end{algorithm}

\section{\label{map_properties}Suboptimality bounds for state and unknown disturbance evolution tracking}
The expansion of RVI for tracking targets with unknown dynamics and the derivation of suboptimality bounds is our main focus. In our approach we solve (\ref{problem_reduce}) approximately with Algorithm \ref{RVI_pseudocode} to find a control sequence $\sigma^{\epsilon,\delta} \in \mathcal{U}^N$ without optimization for unknown input evolution tracking. The suboptimality of state evolution tracking incurred by pruning nodes can then be upper bounded via a worst case analysis as in \cite{Atanasov2014}. We then leverage the relationship between state and unknown input estimation error covariance maps to derive corresponding bounds for unknown input evolution tracking. To begin, we require the following  assumptions.
\begin{assumption}\cite{Atanasov2014}\label{motion_cty}
The sensor motion model is Lipschitz continuous in $x$ with Lipschitz constant $L_f \geq 0$ for every fixed $u \in U$, i.e. $d_\mathcal{X}(f(x_1,u), f(x_2, u)) \leq L_f d_\mathcal{X}(x_1, x_2)$.
\end{assumption}
\begin{assumption}\cite{Atanasov2014}\label{info_cty}
For any two nodes $(x^1_{k-1},\Sigma_{k-1},\Sigma^d_{k-2})$, $(x^2_{k-1},\Sigma_{k-1},\Sigma^d_{k-2})$. Let $\Sigma_{k}^1$, $\Sigma_{k}^2$ be the updated state estimation error covariances after applying control $u \in \mathcal{U}$ to each node. Then
    \begin{equation*}
    \begin{array}{ll}
        \Sigma_k^1 \succeq \gamma\Sigma_k^2 + (1-\gamma)Q_{k-1}, \\
        \Sigma_k^2 \succeq \gamma\Sigma_k^1 + (1-\gamma)Q_{k-1},
    \end{array}
    \end{equation*}
    $\forall k \in [1,N]$, where $\gamma = (1+L_md_{\mathcal{X}}(x_{k}^1, x_k^2))^{-1} < 1$ for some $L_m > 0$. Note for some $\delta > 0$, if $d_{\mathcal{X}}(x_{k-1}^1, x_{k-1}^2) < \delta$ then $\gamma = (1+L_mL_f\delta)^{-1} < 1$.
\end{assumption}


\subsection{Suboptimality bounds for target state evolution tracking\label{mp_state}}

The following properties of the state estimation covariance update map in (\ref{neat_update}) are key for performance analysis.
\begin{lemma}
\label{mc_theorem}The state estimation covariance update map is:
    \begin{enumerate}
        \item Monotone: if $\Sigma_1 \preceq \Sigma_2$ then $\rho(\Sigma_1) \preceq \rho(\Sigma_2)$
        \item Concave: $\forall \alpha \in [0,1]$, $\rho(\alpha\Sigma_1 + (1-\alpha)\Sigma_2) \succeq \alpha\rho(\Sigma_1) + (1-\alpha)\rho(\Sigma_2)$
    \end{enumerate}
\end{lemma}

It is important in our worst case analysis to consider recursive update of the error covariance over a long horizon. We therefore introduce the k-horizon mapping \cite{Tomlin2012} $\phi_{k}:\Sigma_0 \mapsto \Sigma_k$, which maps
the state error covariance matrix at time 0 to time $k$
according to the first $k$ elements $u_{0},\ldots ,u_{k-1}$ of the
control sequence $\sigma \in \mathcal{U}^{N}$:%
\begin{equation}\label{k_map}
\phi^{\sigma}_{k}(\Sigma _{0})=\rho _{u_{k-1}}(\ldots \rho _{u_{1}}(\rho
_{u_{0}}(\Sigma _{0})))=\Sigma _{k}.
\end{equation}%

Monotonicity and concavity of the
k-horizon mapping naturally follow from Lemma \ref{mc_theorem} and the definition in (\ref{k_map}). As a direct result of concavity, the
k-horizon mapping $\phi _{k}$ is bounded by its first order Taylor
approximation, i.e.
\begin{equation}
\phi^{\sigma}_{k}(\Sigma +\epsilon X)\preceq \phi^{\sigma}_{k}(\Sigma )+\epsilon
g^{\sigma}_{k}(\Sigma ,X),  \label{Taylor}
\end{equation}%
where
\begin{equation*}
g^{\sigma}_{k}(\Sigma ,X)=\left. \frac{d\phi^{\sigma} _{k}(\Sigma +\epsilon X)}{d\epsilon }%
\right\vert _{\epsilon =0}
\end{equation*}%
is the directional derivative of the k-horizon mapping $\phi _{k}$ at $\Sigma
\succeq 0$ along an arbitrary direction $X \succeq 0$. The directional derivative $g^{\sigma}_{k}(\Sigma,X)$ can therefore be interpreted as the impact an early perturbative error will have on the error covariance at a later time $k$ provided no further perturbations occur. This interpretation becomes pertinent for studying the consequences of pruning nodes if one frames the $\epsilon I$ term in Definition \ref{alg_redun} as a perturbative error. This motivates the study of the directional derivative.

\begin{lemma}
\label{dir_deriv_single} The directional derivative of the state estimation
covariance update map at $\Sigma \succeq 0$ along the arbitrary
direction $X \succeq 0$ is given by%
\begin{equation*}
\left. \frac{d\rho_{u}(\Sigma +\epsilon X)}{d\epsilon }\right\vert _{\epsilon
=0}=\widetilde{A}(\Sigma )X\widetilde{A}(\Sigma )^{\mathrm{T}},
\end{equation*}%
where $\widetilde{A}(\Sigma )$ is defined as in (\ref{neat_update}). The directional derivative of the
k-horizon mapping $\phi _{k}$ at $\Sigma \in \mathcal{A}$ along an arbitrary
direction $X\in \mathcal{A}$ is given by%
\begin{equation*}\label{dir_deriv_state}
g^{\sigma}_{k}(\Sigma ,X)=\prod_{t=0}^{k-1}(\widetilde{A}_{k-t})X\prod_{t=0}^{k-1}(\widetilde{%
A}_{t})^{\mathrm{T}},
\end{equation*}%
$\forall k=1,\ldots ,N$, with $g^{\sigma}_{0}(\Sigma
,X)=X$.
\end{lemma}


\begin{lemma}\label{dd_decay}
    Suppose $\exists \beta < \infty$ such that $\Sigma_k \preceq \beta I$ $\forall k \in [0,N]$, then we have
    \begin{equation*}
        \text{Tr}\{g^{\sigma}_k(\Sigma,X)\} \leq \beta\eta^k\text{Tr}\{\Sigma^{-1}X\}
    \end{equation*}
    where $\eta = \frac{\beta}{\beta + \underline{\lambda}_Q } < 1$ and $\underline{\lambda}_Q$ is the minimum eigenvalue of $\widetilde{F}_kQ_{k-1}\widetilde{F}_k^{\mathrm{T}}$ $\forall k \in [0,N]$.
\end{lemma}

\indent As in \cite{Atanasov2014}, \cite{Tomlin2012}, the above bound implies that provided the state error covariance is bounded for all time, the effect of a perturbation at an early time step \textit{decays exponentially} as time evolves. The culmination of utilising the above results in a worst case performance analysis is an upper bound on the suboptimality of the state error covariance $\Sigma_N^{\epsilon,\delta}$ found by Algorithm \ref{RVI_pseudocode}. Denoting $J(\cdot) := \log\det(\cdot)$,

\begin{theorem} \label{bounds}
Let $\beta^* < \infty$ be the peak state estimation error of the optimal trajectory, i.e. $\Sigma_k^* \preceq \beta^* I$ $\forall k \in [1,N]$. Then we have
\begin{align*}
0 \leq J(\Sigma_N^{\epsilon,\delta}) - J(\Sigma_N^*) &\leq (\zeta_N - 1)\left(J(\Sigma_N^*) - J(\underline{\lambda}_Q I)\right) + \\&\epsilon(\frac{n_y}{\underline{\lambda}_Q} + \Delta_N)%
\end{align*}
where $\zeta_k 
:= \prod_{\tau=1}^{k-1}\left(1 + \sum_{s=1}^{\tau}L_f^s L_m \delta \right) \geq 1$, 
$\Delta_N := \frac{n_y}{\underline{\lambda}^2_Q}\beta^*\sum_{\tau=1}^{N-1}\frac{\zeta_N}{\zeta_\tau}\eta_*^{N-\tau}$,
$\eta_* = \frac{\beta^*}{\beta^* + \underline{\lambda}_Q} < 1$.
\end{theorem}

This bound is similar to the state estimation bound in \cite{Atanasov2014}, but derived for time varying $Q_k$ using the map in (\ref{neat_update}). As in \cite{Atanasov2014}, \cite{Tomlin2012}, the performance bound in Theorem \ref{bounds} grows with $\delta$ and $\epsilon$, the tunable parameters that dictate pruning. For $\epsilon, \delta = 0$ we recover the optimal solution. 

\subsection{Suboptimality bounds for unknown input tracking\label{mp_input}}

In this section, we show that despite considering only minimisation of the cost function for state estimation as written in (\ref{problem_reduce}), one can still derive concrete suboptimality bounds for the resulting unknown input evolution tracking. 

Once again we introduce a ``k-horizon" update map for unknown input estimation error,
$\phi^d_{k}:\Sigma_0\mapsto\Sigma_{k-1}^d$
\begin{equation}\label{k_ho_input}
\phi^{d\sigma}_{k}(\Sigma _{0})=\rho^d(\phi^{\sigma}_{k-1}(\Sigma_0)) = (F_k^{\mathrm{T}}\widetilde{R}_{k}^{-1}(\Sigma
_{k-1})F_{k})^{-1},
\end{equation}
where $\phi^{\sigma}_{k-1}$ is as in (\ref{k_map}).
From this definition, we see that the control sequence $\sigma^* \in \mathcal{U}^N$ that solves the reduced problem (\ref{problem_reduce}) which considers state error covariance only should give $\Sigma_{N-1}^*$ that minimises (\ref{k_ho_input}). The performance of unknown input tracking should therefore be closely linked to that of the state evolution tracking. However, as $F_N(x_N) = C_N(x_N)G_{N-1}$ the sensor state $x_{N}^*$ found by solving (\ref{problem_reduce}) may not coincide with the sensor state $x_N^{d*}$ required to minimise (\ref{k_ho_input}) over both arguments $\Sigma_{N-1}, x_{N}$. This is an important observation that has direct impact on the performance of unknown input tracking under a control sequence tailored for state evolution tracking optimization. This impact will become apparent in Theorem \ref{input_subopt}.

Monotonicity and concavity of the unknown input error covariance update map are again crucial properties for describing the evolution of nodes.

\begin{lemma}
\label{input_map}The unknown input error covariance update map $\rho ^{d}(\cdot )$
is monotone and concave.
\end{lemma}

Lemma \ref{input_map} extends to the k-horizon input estimation error update map in (\ref{k_ho_input}). Thus, $\phi^{d\sigma}_{k}$ is bounded from above by its first order Taylor approximation. We can again characterize the directional derivative $g_{k-1}^{d\sigma}(\Sigma,X)$. 

\begin{lemma}
\label{dir_deriv_input} The directional derivative of $\phi^{d\sigma}_{k}$  at $\Sigma \succeq 0 $ in the direction $X\succeq 0 $ is given by%
\begin{align*}
g_{k-1}^{d\sigma}(\Sigma,X) &= \left. \frac{d}{d\epsilon}\phi _{k-1}^{d}(\Sigma +\epsilon X)\right\vert
_{\epsilon =0}\\
&=M_{k}^*C_{k}A_{k-1}g^{\sigma}_{k-1}(\Sigma ,X)A_{k-1}^{\mathrm{T}}C_{k}^{%
\mathrm{T}}M_{k}^{*\mathrm{T}},
\end{align*}%
where $g^{\sigma}_{k-1}(\Sigma ,X)$ is the directional derivative of the state k-horizon update map.
\end{lemma}

As in Lemma \ref{dd_decay}, we find that the effect of a perturbation in the state error covariance on the unknown input error covariance dampens with time provided $\Sigma_k^d$ and $\Sigma_k$ are bounded for all $k$. 
\begin{lemma}\label{input_dd_decay} Suppose $\exists \beta^d < \infty$ such that $\Sigma_k^d \preceq \beta^d I$ $\forall k \in [1,N]$. Then 
\begin{equation*}
    \text{Tr}\{g_{k-1}^{d\sigma}(\Sigma, I)\} \leq (n_d)^2(\beta^{d})^2\overline{\lambda}_{\widetilde{G}}\text{Tr}\{g^{\sigma}_{k-1}(\Sigma, I)\}
\end{equation*}
where $\overline{\lambda}_{\widetilde{G}}$ is the maximum eigenvalue of $G_{k-1}^{\mathrm{T}}H_kA_{k-1}A_{k-1}^{\mathrm{T}}H_kG_{k-1} \in \mathbf{R}^{n_d\times n_d}$.
\end{lemma}

The propagated error incurred on the unknown input error covariance by a perturbation in the state covariance is therefore a multiple of that found for the state. Hence, given the state result in Lemma \ref{dd_decay}, the unknown input analogue can also be found. We now provide an upper bound on the final unknown input error covariance found by Algorithm \ref{RVI_pseudocode}.
\begin{theorem} \label{input_subopt}
Let $\beta^* < \infty$, $\beta^{d*} < \infty$ be the peak state and input estimation errors of the optimal trajectory respectively. That is, $\Sigma_k^* \preceq \beta^* I$ and $\Sigma_{k-1}^{d*} \preceq \beta^{d*} I$ $\forall k \in [1,N]$. Then
\begin{equation*}
\begin{array}{l}
0 \leq J(\Sigma_{N-1}^{d,(\epsilon,\delta)}) - J(\Sigma_{N-1}^{d*}) \\\leq (\zeta_N - 1)\left(J(\Sigma_{N-1}^{d*})+J(\gamma^{d*} I ) - J(\overline{\lambda}_H^{-1}I)\right) + \epsilon(\Delta_N^d)
\end{array}
\end{equation*}
where $\Delta_N^d := (\gamma^{d*})^{-1}(n_d)^2 \overline{\lambda}_H\overline{\lambda}_{\widetilde{G}}(\beta^{d*})^2\underline{\lambda}_Q\Delta_N$, $\overline{\lambda}_H$ is the maximum eigenvalue of $G_{N-1}^{\mathrm{T}}H_NG_{N-1}$, and $\gamma^{d*} = (1+L_md_{\mathcal{X}}(x_N^{*},x_N^{d*}))^{-1}$.
\end{theorem}

We observe the same behaviours of the unknown input bounds with respect to $\delta, \epsilon, N$ as the state bounds found in Theorem \ref{bounds}. Here $\Delta_N^d$ is a factor of $\Delta_N$ from the state bounds, again highlighting the close relationship between the two bounds. Most notably, we see the previously mentioned impact of unknown input estimation under a control sequence optimized for state estimation; the bound grows with $(\gamma^{d*})^{-1}$. This result is expected -- if the distance $d_{\mathcal{X}}(x_N^{*},x_N^{d*})$ between optimal sensor positions for state estimation and unknown input estimation is large, the performance of unknown input estimation resulting from optimizing only state estimation via Algorithm \ref{RVI_pseudocode} worsens.

\section{\label{sims}Illustrative Simulations}

In this section, we illustrate the theoretical results with a two-dimensional target tracking problem in which the target dynamics is subject to an unknown input signal. Suppose a sensor with state $x_{k}$ defined by its position-velocity vector is mounted on a robot with the dynamic model:
\begin{equation}
    x_{k+1} = f(x_k,u_k) := \begin{pmatrix}x_{k}^1\\ x_{k}^2\\0\\0\end{pmatrix} + \begin{pmatrix}u_{k}^1\cos(u_{k}^2)\tau\\u_{k}^1\sin(u_{k}^2)\tau\\u_{k}^1\cos(u_{k}^2)\\u_{k}^1\sin(u_{k}^2)\end{pmatrix}
\end{equation}
with control input $u_k \in \mathcal{U}$, where $\mathcal{U} = \{(u_{k}^1, u_{k}^2) \mid u_{k}^1 \in \{0,1,2\}, u_{k}^2 \in \{0, \pm \pi/2, \pi\}$ and $\tau$ is a small time translation. The goal of the robot is to track and estimate the position and velocity of a constant-velocity vehicle driven by Gaussian noise and an unknown input $d_k$ in the form of abrupt accelerations:
\begin{equation}
\begin{array}{l}
    y_{k+1} = \begin{bmatrix}I_2 & \tau I_2\\0 & I_2\end{bmatrix}y_k + \begin{bmatrix}\tau^2/2 I_2 \\ \tau I_2\end{bmatrix}d_k + w_k, \\ w_k \sim \mathcal{N}\left(0, q\begin{bmatrix} \tau^3/3 I_2 & \tau^2/2 I_2\\ \tau^2/2I_2 & \tau I_2\end{bmatrix}\right)
\end{array}
\end{equation}
where $y_k = [y_{k}^1, y_{k}^2, \dot{y}_{k}^1, \dot{y}_{k}^2]^T$ is the position-velocity vector of the target state at time $k$ and $q$ is a diffusion strength scalar. The tracking takes place over 51 time steps. At $k \in \{4, 9, 19, 24, 34, 39\}$, $d_k$ is a maneuver that takes form of a sharp acceleration in some direction.

\begin{figure}[t]
\centering
\includegraphics[width=\columnwidth]{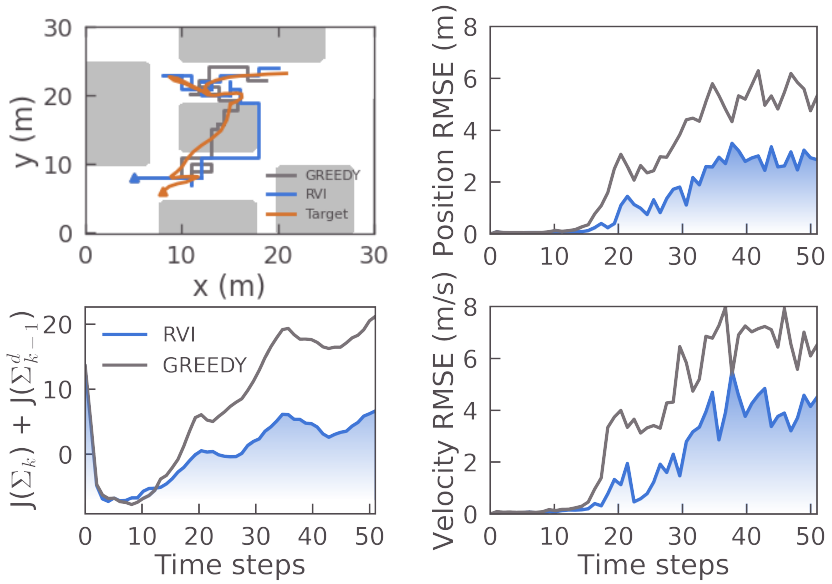}
\caption{\label{fig1}Simulation results of the target tracking problem averaged over 150 MC simulations. The left panel shows an example trajectory (initial robot positions marked by triangles) and the cost of each policy's calculated trajectory. The right panel shows the average RMSE of each policy's target position and velocity estimates.}
\end{figure}

The sensor takes noisy position measurements of the target and uses them to obtain the target's velocity by differentiation. For simplicity, the sensor observation model in (\ref{sensor_mea}) is given by $C_k = I_4$  with the measurement noise increasing linearly with the distance between robot and target. Certain areas of the environment are ``cloudy", depicted as grey areas in Figure \ref{fig1}, and increase the robot's measurement noise. Upon entering a cloud, the robot should slow down for safety under poor visibility. Beyond a maximum range of 20 metres the measurement noise is effectively infinite.

For $150$ Monte-Carlo simulations, Algorithm \ref{RVI_pseudocode} is used to track the target with $\tau = 1$, $N=5$, $q=0.1$, $\epsilon = 0.1$, $\delta = 1$. The performance of our proposed algorithm is compared to a greedy approach in Figure \ref{fig1}. The RVI algorithm's long planning horizon predicts the target will enter and remain in an area of high measurement noise in future time steps, and thus prioritises avoiding entering this area over remaining close to the target. On the contrary, the greedy algorithm prioritises minimising the cost function at each time step and therefore lacks the foresight to avoid these areas.

The trajectory costs of the two policies in Figure \ref{fig1} elucidate the impact that RVI's non-myopic planning has on the performance of the found solution. We see that RVI incurs much less cost than the greedy policy. Further, comparison of the average root mean square error (RMSE) of the policies' state estimates shows that for all time steps our algorithm more successfully tracks target evolution in the presence of unknown inputs. These results are promising confirmation of our theoretical expansion of RVI to tracking targets subject to arbitrary, unknown disturbances.

\section{\label{conclusion}Conclusion and Discussions}

In this work, we studied the AIA problem for targets subject to arbitrary unknown disturbances. We have shown both the state and input error covariance update maps given in existing unknown input filtering works are concave and monotone. These properties were used to derive suboptimality guarantees for both state and unknown disturbance tracking by the proposed method. Notably, we have shown that one may consider tracking only the target state without loss of performance guarantees for unknown disturbance tracking due to the close relationship between unknown disturbance and target state estimation. 
The suboptimality bounds presented were notably linear in the tuning parameters which dictate strictness of node pruning, and thus the optimal solution is recovered when the tuning parameters are set to zero. Simulations demonstrated that the proposed algorithm performs well in tracking a target undertaking unknown maneuvers. Future work will focus on the more general case with unknown disturbances affecting both target motion and sensor observation models. A decentralized extension of the AIA considered here will also be pursued.

\section{Appendix A: Proofs for main results}
\subsection{Proofs for results in Section \ref{mp_state}}

\begin{proof}[Proof of Lemma \ref{mc_theorem}, \ref{dir_deriv_single}, \ref{dd_decay}]
The proofs are algebraically involved and therefore
skipped due to limited space.
\end{proof}

\subsection{Proofs for results in Section \ref{mp_input}}
To prove Lemma \ref{input_map}, we require some preparatory results:
\begin{lemma}
\label{inside_out_arg}Let $\alpha \in \lbrack 0,1]$ be a constant. Then $%
\forall \Sigma \succeq 0 $, we have $\alpha \rho ^{d}(\Sigma )\preceq \rho ^{d}(\alpha \Sigma )$.
\end{lemma}

\begin{proof}[]
For $\Sigma \succeq 0$, denote $\overline{R}_{k}(\Sigma
)=C_{k}A_{k-1}\Sigma A_{k-1}^{\mathrm{T}}C_{k}^{\mathrm{T}}+\alpha
^{-1}C_{k}Q_{k-1}C_{k}^{\mathrm{T}}+\alpha ^{-1}R_{k}$. We have%
\begin{equation*}
\begin{array}{l}
\overline{R}_{k}(\Sigma )-\widetilde{R}_{k}(\Sigma )=(\alpha ^{-1}-1)(C_{k}Q_{k-1}C_{k}^{\prime }+R_{k})\succeq 0,%
\end{array}%
\end{equation*}%
when $\alpha \in (0,1]$ and $R_{k}\succ 0.$ Thus, $\overline{R}_{k}(\Sigma
)\succeq \widetilde{R}_{k}(\Sigma )$. Since $X\mapsto X^{-1}$ is order reversing
for any matrix $X$, we have $\widetilde{R}_{k}^{-1}(\Sigma )\succeq \overline{R}%
_{k}^{-1}(\Sigma )$. Additionally, note that%
\begin{equation*}
\begin{array}{l}
\rho ^{d}(\alpha \Sigma )=(F_{k}^{\prime }\widetilde{R}_{k}^{-1}(\alpha \Sigma
)F_{k})^{-1} \\
=\alpha (F_{k}^{\mathrm{T}}\overline{R}_{k}^{-1}(\Sigma )F_{k})^{-1}\succeq
\alpha (F_{k}^{\mathrm{T}}\widetilde{R}_{k}^{-1}(\Sigma )F_{k})^{-1}=\alpha \rho
^{d}(\Sigma ).%
\end{array}%
\end{equation*}%
For $\alpha =0$, we have $\alpha \rho ^{d}(\Sigma )=0\preceq
\rho ^{d}(\alpha \Sigma )$.  
\end{proof}

\begin{lemma}
\label{convex_orig} $f(\Sigma )=F_{k}^{\mathrm{T}}\widetilde{R}_{k}^{-1}(\Sigma
)F_{k}$ is a convex function of $\Sigma \in \mathcal{A}$.
\end{lemma}

\begin{proof}[]
We note that $\widetilde{R}_{k}(\cdot )$ is monotone. Then, by Corollary V.2.6
in \cite{Bhatia2013}, $\widetilde{R}_{k}^{-1}(\cdot )$ is operator convex.
For $\Sigma _{1},\Sigma _{2} \succeq 0$ and $\alpha \in \lbrack 0,1]$,
let $\chi =\alpha \Sigma _{1}+(1-\alpha )\Sigma _{2}$. We can prove that%
\begin{equation*}
F_{k}^{\mathrm{T}}(\alpha \widetilde{R}_{k}^{-1}(\Sigma _{1})+(1-\alpha )\widetilde{R%
}_{k}^{-1}(\Sigma _{2})-\widetilde{R}_{k}^{-1}(\chi ))F_{k}\succeq 0.
\end{equation*}%
So $f(\Sigma )$ is also operator convex.
\end{proof}

\begin{proof}[Proof of Lemma \ref{input_map}]
We note that $\widetilde{R}_{k}(\cdot )$ is monotone. Then for any $\Sigma_1, \Sigma_2 \succeq 0$ with $\Sigma_1 \preceq \Sigma_2$, we have $\widetilde{R}_{k}(\Sigma_1)\preceq\widetilde{R}_{k}(\Sigma_2)$. Then, as matrix multiplication is order preserving, and matrix inversion is order reversing, it immediately follows that $\rho ^{d}(\Sigma _{1})\preceq \rho ^{d}(\Sigma _{2})$. Hence, monotonicity is proved. We next prove concavity. For $\forall \alpha
\in \lbrack 0,1]$ and $\forall \Sigma _{1},\Sigma _{2}\succeq 0$, let $%
\chi =\alpha \Sigma _{1}+(1-\alpha )\Sigma _{2}$. Then, from Lemma \ref{convex_orig}, and since $\alpha \widetilde{R}_{k}^{-1}(\Sigma )\preceq \widetilde{R}_{k}^{-1}(\alpha \Sigma )$ we have $F_{k}^{\mathrm{T}}\widetilde{R}_{k}^{-1}(\chi )F_{k}\preceq F_{k}^{\mathrm{T}}\widetilde{R}_{k}^{-1}(\alpha \Sigma _{1})F_{k}+F_{k}^{%
\mathrm{T}}\widetilde{R}_{k}^{-1}((1-\alpha )\Sigma _{2})F_{k}$. Inverting this expression, utilising Lemma \ref{inside_out_arg}, and remembering that $X\mapsto X^{-1}$ is a convex operation \cite{Bhatia2013} gives
\begin{align*}
&\rho ^{d}(\chi )-\alpha \rho ^{d}(\Sigma _{1})-(1-\alpha )\rho ^{d}(\Sigma
_{2})\\
&\succeq \lbrack F_{k}^{\mathrm{T}}\widetilde{R}_{k}^{-1}(\chi )F_{k}]^{-1} \\&-[F_{k}^{\mathrm{T}}\widetilde{R}_{k}^{-1}(\alpha \Sigma _{1})F_{k}+F_{k}^{%
\mathrm{T}}\widetilde{R}_{k}^{-1}((1-\alpha )\Sigma _{2})F_{k}]^{-1}\succeq 0,
\end{align*}
thus proving concavity.
\end{proof}

\begin{proof}[Proof of Lemma \ref{dir_deriv_input}]
Denoting $V = \left. \frac{d}{d\epsilon}\widetilde{R}_{k}^{-1}(\phi^{\sigma} _{k-1}(\Sigma +\epsilon X))\right\vert _{\epsilon =0}$ and $\widetilde{R}%
_{k}^{-1}(\phi^{\sigma}_{k-1}(\Sigma )) = \widetilde{R}_k^{-1}$, it is simple to show that
\begin{equation*}
\begin{array}{l}
 \left. \frac{d}{d\epsilon}\phi _{k-1}^{d\sigma}(\Sigma +\epsilon X)
\right\vert _{\epsilon =0} 
=-\phi _{k-1}^{d\sigma}(\Sigma )F_{k}^{\mathrm{T}}VF_{k}\phi _{k-1}^{d\sigma}(\Sigma ),\\
V=-
\widetilde{R}_{k}^{-1} 
\left. \frac{d}{d\epsilon}\widetilde{R}_{k}(\phi ^{\sigma}_{k-1}(\Sigma +\epsilon X))\right\vert _{\epsilon =0}\widetilde{R}_{k}^{-1}.%
\end{array}
\end{equation*}%
Putting these together and simplifying gives the result.
\end{proof}

\begin{proof}[Proof of Lemma \ref{input_dd_decay}]
The proof follows straightforwardly by considering the cyclical property of trace operator and submultiplicity of the Frobenius norm $\vert\vert\cdot\vert\vert = \sqrt{\text{Tr}\{\cdot\}}$.
\end{proof}
\subsection{Proof of Theorem \ref{bounds}}

\begin{proof}[Proof of Theorem \ref{bounds}]
The proof follows from applying $J(\cdot)$ to Appendix C Lemma 7 of ~\cite{Atanasov2014} to calculate the bounds, noting that $\phi_{k-1-\tau}^{\sigma_{\tau}^*}(Q_{\tau}) \succeq \widetilde{F}_{\tau+k-1}Q_{\tau+k-2}\widetilde{F}_{\tau+k-1}^{\mathrm{T}}$ $\forall k, \tau$ and $\widetilde{F}_{k}Q_{k-1}\widetilde{F}_{k}^{\mathrm{T}} \succeq \underline{\lambda}_Q I$ $\forall k$. 
\end{proof}

\subsection{Proof of Theorem \ref{input_subopt}}
\begin{lemma}\label{h_lemma}
There exists a real constant $L_m \geq 0$ such that $\forall x_1, x_2 \in \mathcal{X}$:%
\begin{equation*}
\begin{array}{l}
H_k(x_1) \preceq (1 + L_m d_\mathcal{X}(x_1, x_2))H_k(x_2),\\
H_k(x_2) \preceq (1 + L_m d_\mathcal{X}(x_1, x_2))H_k(x_1)
\end{array}
\end{equation*}%
where $H_k(x) = C_k^{\mathrm{T}}(x)R_k^{-1}(x)C_k(x)$.
\end{lemma}

\begin{proof}[]
Consider any two nodes $(x^1_{k-1},\Sigma_{k-1},\Sigma^d_{k-2})$, $(x^2_{k-1},\Sigma_{k-1},\Sigma^d_{k-2})$. Then, applying control $u \in \mathcal{U}$ to each node we have
$\rho_{x^1_{k}}(\Sigma_{k-1}) \succeq \gamma\rho_{x^2_{k}}(\Sigma_{k-1})$ from Assumption \ref{info_cty}.
Hence,
\begin{equation}\label{inverse_relo}
\gamma^{-1}\rho_{x^2_{k}}^{-1}(\Sigma_{k-1}) \succeq \rho_{x^1_{k}}^{-1}(\Sigma_{k-1}) \succeq \rho_{x^1_{k}}^{-1}(\gamma^{-1}\Sigma_{k-1}),   
\end{equation} where the last inequality follows from monotonicity of $\rho$ and $\gamma^{-1} > 1$.
Denote
\begin{align*}
     \overline{\Sigma}^{-1}_{k+1}(\Sigma_k) &= A_{k}^{-T}\Sigma_{k}^{-1}A_{k}^{-1} \\ - A_{k}^{-T}&\Sigma_{k}^{-1}A_{k}^{-1}(A_{k}^{-T}\Sigma_{k}^{-1}A_{k}^{-1} + Q_{k}^{-1})^{-1}A_{k}^{-T}\Sigma_{k}^{-1}A_{k}^{-1},
\end{align*}
then it is simple to show \begin{equation}\label{bar_result}\gamma\overline{\Sigma}_{k+1}^{-1} \preceq (\overline{\gamma^{-1}\Sigma_{k+1}})^{-1}.\end{equation}

Now, the information form covariance update map for the filter in (\ref{pre_without})-(\ref{definition_error}) is $\rho^{-1}(\cdot)$ \cite{Gillijns2007B}: 
\begin{equation*}
\Sigma _{k}^{-1}=\overline{\Sigma }_{k}^{-1}+H_{k}-\overline{\Sigma }%
_{k}^{-1}G_{k-1}(G_{k-1}^{\mathrm{T}}\overline{\Sigma }%
_{k}^{-1}G_{k-1})^{-1}G_{k-1}^{\mathrm{T}}\overline{\Sigma }_{k}^{-1}.
\end{equation*}%
Using (\ref{inverse_relo}) and (\ref{bar_result}) and denoting $%
H_{k}(x_{k}^{n})=H_{k}^{n}$ for $n=1,2$, it follows that
\begin{align*}
& \gamma ^{-1}(\overline{\Sigma }_{k}^{-1}+H_{k}^{2}-\overline{\Sigma }%
_{k}^{-1}G_{k-1}(G_{k-1}^{\mathrm{T}}\overline{\Sigma }%
_{k}^{-1}G_{k-1})^{-1}G_{k-1}^{\mathrm{T}}\overline{\Sigma }_{k}^{-1}) \\
& \succeq \gamma ^{-1}\overline{\Sigma }_{k}^{-1}+H_{k}^{1} \\
& -\gamma ^{-1}\overline{\Sigma }_{k}^{-1}G_{k-1}(G_{k-1}^{\mathrm{T}}%
\overline{\Sigma }_{k}^{-1}G_{k-1})^{-1}G_{k-1}^{\mathrm{T}}\overline{\Sigma
}_{k}^{-1}.
\end{align*}
Reducing gives the desired result $H_k^1 \preceq (1+L_md_{\mathcal{X}}(x_k^1,x_k^2))H_k^2$. Following identical working for $H_k(x_k^2)$ completes the proof. 
\end{proof}

\begin{lemma}\label{input_cov_cty}
    Suppose $(x^1_{k-1},\Sigma_{k-1},\Sigma_{k-2}^d)$, $(x^2_{k-1}, \Sigma_{k-1}, \Sigma_{k-2}^d)$ are two nodes with $d(x^1,x^2) \leq \delta$. Let $\Sigma_{k-1}^{d,1}$, $\Sigma_{k-1}^{d,2}$ be the input estimation error covariances after updating both nodes under the control $u \in \mathcal{U}$ Then
    \begin{equation*}
    \begin{array}{ll}
        \Sigma_{k-1}^{d,1} \succeq \gamma\Sigma_{k-1}^{d,2}, & \Sigma_{k-1}^{d,2} \succeq \gamma\Sigma_{k-1}^{d,1}
    \end{array}
    \end{equation*}
    $\forall k \in [1,N]$.
\end{lemma}

\begin{proof}[] Denote $R_k(x_i) = R_k^i$ for $i = 1,2$, and consider the inverse of the update map applied to node $(x^1_{k-1},\Sigma_{k-1},\Sigma_{k-2}^d)$:
\begin{align*}
&(\rho_{x^1}^d(\Sigma_{k-1}))^{-1} = (\Sigma_{k-1}^{d,1})^{-1} = F_k^{\mathrm{T}}\widetilde{R}_k^{-1}(\Sigma_{k-1})F_k\\
&= F_k^{\mathrm{T}}\left(C_kA_{k-1}\Sigma_{k-1} A_{k-1}^{\mathrm{T}}C_k^{\mathrm{T}} + C_kQ_{k-1}C_k^{\mathrm{T}} + R_k^1\right)^{-1}F_k.
\end{align*}
By applying the matrix inversion lemma and Lemma \ref{h_lemma}, we get $(\Sigma_{k-1}^{d,1})^{-1}\preceq \gamma^{-1}(\rho_{x^2}^d(\gamma^{-1}\Sigma_{k-1}))^{-1}$. Then, taking the inverse and applying monotonicity of the update map gives the result. Following the same reasoning for updating $(x^2_{k-1},\Sigma_{k-1},\Sigma_{k-2}^d)$ completes the proof.
\end{proof}

\begin{proof}[Proof of Theorem \ref{input_subopt}]
Applying $\rho_{u_{N-1}^*}^d(\cdot)$ to Appendix C Lemma 7 of ~\cite{Atanasov2014}, where $u_{N-1}^*$ is the state-optimized control found by the algorithm, rather than an input-optimized control $u_{N-1}^{d*}$. Noting again that $\phi_{k-1-\tau}^{\sigma_{\tau}^*}(Q_{\tau}) \succeq \widetilde{F}_{\tau+k-1}Q_{\tau+k-2}\widetilde{F}_{\tau+k-1}^{\mathrm{T}}$ $\forall k$, and that $\sum_{\tau=1}^{N-1}\Gamma_{\tau}(1-\gamma_{\tau}) = 1-\Gamma_{N-1}$, by concavity of $\rho^d$,
\begin{align*}
    &\rho_{u_{N-1}^*}^d(\Sigma^*_{N-1}) + \\&\hspace{5mm}\epsilon g_1^{d\sigma_{N-1}^*}(\Sigma_{N-1}^*,\sum_{\tau=1}^{N-2}\Gamma_{\tau} g^{\sigma^*_{\tau}}_{N-1-\tau}(\Sigma_{\tau}^*,I) +\Gamma_{N-1} I) \\
    &\succeq \Gamma_{N-1}\sum_{i=1}^K\alpha_i\rho_{u_{N-1}^*}^d(\Sigma_{N-1}^i) + (1-\Gamma_{N-1})\rho_{u_{N-1}^*}^d(\underline{\lambda}_Q I).
\end{align*}
Using Lemma \ref{input_cov_cty} with $\gamma^{d*} = (1+ L_md_{\mathcal{X}}(x_{N}^*, x_{N}^{d*}))^{-1}$, and $\rho_{u_{N-1}^*}^d(\underline{\lambda}_Q I) \succeq (G_{N-1}^{\mathrm{T}}H_NG_{N-1})^{-1}\succeq \overline{\lambda}_H^{-1} I$, we have
\begin{align*}
    &\gamma^{d*}\Sigma^{d*}_{N-1} + M_NC_NA_{N-1}\sum_{\tau=1}^{N-1}\Gamma_{\tau}g_{N-\tau}^{\sigma^*_{\tau}}(\Sigma_{\tau}^*,I)A_{N-1}^{\mathrm{T}}C_N^{\mathrm{T}}M_N^{\mathrm{T}}
    \\&\succeq \Gamma_{N-1}\sum_{i=1}^K\alpha_i\gamma^{d*}\Sigma_{N-1}^{di} + (1-\Gamma_{N-1})\overline{\lambda}_H^{-1} I.
\end{align*}

The proof then follows using monotonicity and concavity of $J(\cdot)$, and Lemma \ref{input_dd_decay} after applying $J(\cdot)$ to the above result. 
\end{proof}

\end{document}